\documentclass{article}

\usepackage{microtype}
\usepackage{graphicx}
\usepackage[dvipsnames]{xcolor}
\usepackage{subfigure}
\usepackage{booktabs} 
\usepackage{amsmath,amssymb,amsfonts}

\usepackage{hyperref}

\usepackage[accepted]{icml2021}

\icmltitlerunning{Integrating LSTMs and GNNs for COVID-19 Forecasting}

\begin{document}

\twocolumn[
\icmltitle{Integrating LSTMs and GNNs for COVID-19 Forecasting}



\icmlsetsymbol{equal}{*}

\begin{icmlauthorlist}
\icmlauthor{Nathan Sesti}{equal,mit}
\icmlauthor{Juan Jose Garau-Luis}{equal,mit}
\icmlauthor{Edward Crawley}{mit}
\icmlauthor{Bruce Cameron}{mit}
\end{icmlauthorlist}

\icmlaffiliation{mit}{Engineering Systems Laboratory, Massachusetts Institute of Technology, Cambridge, MA, USA}
\icmlcorrespondingauthor{Juan Jose Garau-Luis}{garau@mit.edu}

\icmlkeywords{COVID-19, Graph Neural Networks, LSTM, policy-making}

\vskip 0.3in
]



\printAffiliationsAndNotice{\icmlEqualContribution} 


\begin{abstract}

The spread of COVID-19 has coincided with the rise of Graph Neural Networks (GNNs), leading to several studies proposing their use to better forecast the evolution of the pandemic. Many such models also include Long Short Term Memory (LSTM) networks, a common tool for time series forecasting. In this work, we further investigate the integration of these two methods by implementing GNNs within the gates of an LSTM and exploiting spatial information. In addition, we introduce a skip connection which proves critical to jointly capture the spatial and temporal patterns in the data. We validate our daily COVID-19 new cases forecast model on data of 37 European nations for the last 472 days and show superior performance compared to state-of-the-art graph time series models based on mean absolute scaled error (MASE). This area of research has important applications to policy-making and we analyze its potential for pandemic resource control.

\end{abstract}

\section{Introduction}
\label{introduction}

Since its outbreak in late 2019, the COVID-19 virus \cite{Fauci2020,velavan2020covid} has devastated the world, causing over 3 million deaths \cite{WorldHealthOrganization} and economic losses estimated up to \$10 trillion \cite{UnitedNations}. Crucial to curbing such damage is prompt response and effective policy-making, and, even with vaccination numbers increasing, this ability will be crucial to develop as the possibility of a future pandemic looms. Due to the exponential nature of epidemiological transmission \cite{Li2020}, even slight improvements on early intervention can have outsized impact, making preemptiveness one of our best tools. Therefore, knowledge of future spread is of critical importance.

Given the urgency of the pandemic, the Machine Learning (ML) community has stepped in to help in multiple capacities \cite{Alimadadi2020,Sills145}, from assisting in early COVID-19 diagnosis based on CT scans \cite{Barstugan2020,WANG2021208}, to predicting new cases and hospitalizations using different time series models \cite{alazab2020covid}. Among the latter efforts, some authors have taken advantage of both the temporal correlations and the inherent network structure of epidemiological data and thus combined temporal models such as Long Short Term Memory (LSTM) networks \cite{hochreiter1997long} with recent advancements in Graph Neural Networks (GNNs) \cite{scarselli2008graph}. This has proven to be one of the most successful approaches.

In this work we aim to improve upon their accuracy, presenting a new time series forecasting model based on GNNs and LSTMs. We introduce a general method that further integrates both networks and apply it to the problem of forecasting new COVID-19 cases. Our specific use case consists of 37 countries in Europe, with data from Jan. 2020 to May 2021. We show that our model is able to forecast new cases out of the training distribution, showing better results than four other GNN+LSTM models and lag, with a prediction error of 10\% when forecasting 7 days ahead. We also analyze how the model performance translates into real-world policy-making by means of the fraction of missed cases. Finally, our findings suggest that introducing a skip connection in our model is key to reducing underfitting.

\begin{figure*}[t]
\centering
\includegraphics[width=.99\linewidth]{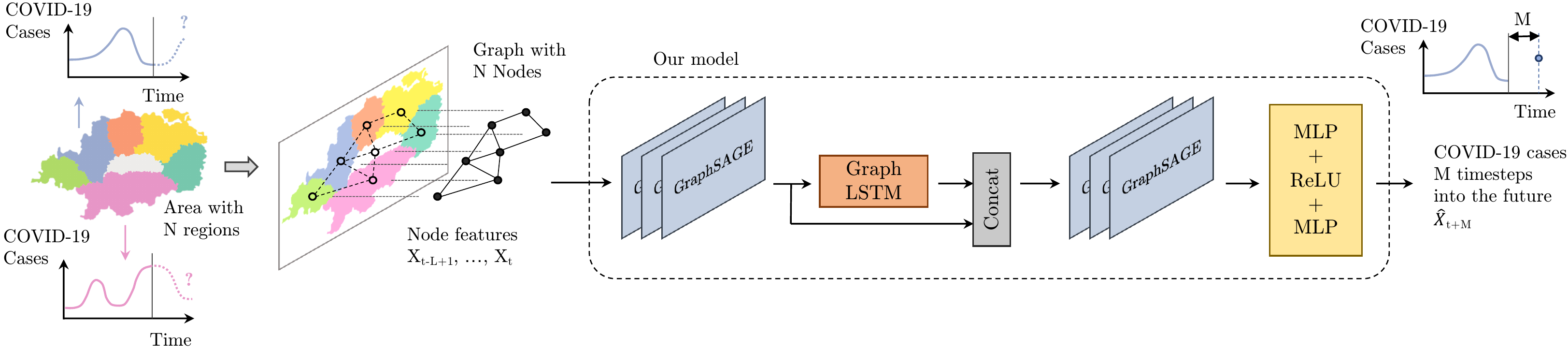}
\caption{Overview of the problem and our proposed method. We consider an area with $N$ different regions (e.g., different countries, states), each displaying a different COVID-19 case curve. The task consists of predicting, $M$ timesteps in advance (we use days in our paper), the number of cases in each region. To that end, a graph with $N$ nodes is created based on the geographical structure of the regions and a feature vector is associated to each node. These are fed to our model, consisting of three layers of GraphSAGE convolutions, an LSTM with graph convolution gates, three additional GraphSAGE layers, and two final fully-connected layers. We use a skip connection between the first batch of GraphSAGE layers and the GraphLSTM.}
\label{fig:arch}
\end{figure*}

\section{Related Work}


Following the outbreak of COVID-19, the ML community has given the problem of epidemiological modeling a refreshed solution. The majority of the studies have made LSTMs the tool of choice. In addition, many researchers have answered the call for use of mobility data \cite{Sills145,santosh2020} and used graphs to model spatial information, some converging upon different combinations of LSTMs and GNNs to solve the problem, mirroring their use in other applications.

\textbf{Temporal-Only Models}\,\,\, Since early in the pandemic, researchers have realized the value of LSTM networks due to the salient temporal patterns of transmission \cite{chimmula2020, shahid2020, arora2020}. Their predictions displayed low error, competing well with standard epidemiological models, such as susceptible-infected-recovered-dead (SIRD) models \cite{bailey1975mathematical}.

\textbf{GNNs for COVID-19}\,\,\, Capitalizing on the call for use of mobility data, numerous papers have applied GNNs to the task. Some have leveraged classic epidemiological models, one feeding the concatenation of outputs from a GNN and an epidemiological model into a final distributional regression layer \cite{LaGatta2021}. Alternatively, \cite{fritz2021} uses a combination of a GNN and LSTM to predict the parameters of a SIRD model. Our work will differ in that it is not dependent upon any explicit epidemiological model, instead being able to construct its own parametrization of input features.

While many studies take advantage of both the spatial and temporal dependencies of the problem, \cite{cao2021} pays particular attention to modeling them jointly --- i.e., leveraging the correlations that depend on both other regions and timesteps rather than taking these separately. While they achieve this using Fourier Transforms, we instead draw from \cite{seo2018}, which replaces the typical linear transformation within an LSTM with a Chebyshev Spectral CNN \cite{defferrard2016}, although we switch it out with GraphSAGE \cite{hamilton2017}. Our contribution also includes a novel skip connection, which concatenates the output of the aforementioned LSTM module with a purely spatial GNN in order to improve upon the challenges common to joint modeling of spatial and temporal dimensions.

\textbf{Other Applications of GNN + LSTM}\,\,\, At the intersection of these two network models has been an explosion of research aiming to tackle other graph-structured time series problems. Specifically, DCRNN \cite{li2017} was applied to traffic prediction, GCLSTM \cite{chen2018} was applied to predicting emails and other contact between people, and GConvGRU \cite{seo2018} was able to predict moving MNIST data and model natural language.


\section{Problem Statement}

Our objective in this paper is to predict some numerical feature of each node in a graph given a sequence of previous graph snapshots. In the case of COVID-19, we want to predict new cases across different regions, where each region constitutes a different node in the graph. The data can be characterized as a graph $\mathcal{G}=(X,A,W)$ where $X=\{x_{ijt}\}\in \mathbb{R}^{K_X\times N\times T}$ indicates the $K_X$ input features for each of $N$ nodes at $T$ different timesteps, with $X_t\in \mathbb{R}^{K_X\times N}$ indicating the values at timestep $t$. Then, $A$ is the adjacency matrix ---fixed over time--- and $W\in \mathbb{R}^{K_W \times N\times N}$ represents each of the $K_W$ static edge features used in our model between each pair of countries.

The task at timestep $t$ is, given the $L$ most recent timesteps $\{X_{t-L+1},\mathellipsis,X_{t}\}
$, to predict the number of new cases $M$ timesteps into the future, $\hat{X}_{t+M}\in \mathbb{R}^{N}$. When each graph snapshot corresponds to a unique day, the task becomes predicting new cases $M$ days ahead, which benefits effective policy-making and early response if $M$ is large enough.


\section{Methods}

Our method consists of a combination of GraphSAGE and LSTM layers that jointly exploit spatio-temporal relationships in the data. This section covers the main details of the model, wich is depicted in Figure \ref{fig:arch}. Our approach considers edge features in the GraphSAGE layers and replaces linear operations in the LSTM by graph convolutions.

\textbf{GraphSAGE}\,\,\, Our model makes extensive use of GraphSAGE \cite{hamilton2017}, a spatial GNN that aggregates the features of local nodes to generate an embedding for each node in a graph, $\mathcal{G} = (\mathcal{V}, \mathcal{E})$. Let $\boldsymbol{x}_j=\boldsymbol{h}_j^0$ represent the feature vector of node $j$, which is also the embedding for that node before any iterations. In each iteration $k\in \{1,...,n\}$, each node $v \in \mathcal{V}$ aggregates the embeddings of all immediately adjacent nodes $\{\boldsymbol{h}_u^{k-1}, \ \forall u \in \mathcal{N}(v)\}$, usually by taking the mean of these vectors. This aggregated vector, $\boldsymbol{h}_{\mathcal{N}(v)}^{k-1}$, is then concatenated to the current embedding $\boldsymbol{h}_v^{k-1}$ and fed into a single neural network layer with a sigmoid activation function to compute $\boldsymbol{h}_v^k$. We slightly alter this scheme to allow for edge features $e_{vu}$ between any two nodes $u$ and $v$, taking a weighted mean, $\boldsymbol{h}_{\mathcal{N}(v)}^{k-1} = \frac{1}{|\mathcal{N}(v)|}\sum\limits_{u\in\mathcal{N}(v)}{e_{vu}\cdot h_u^{k-1}}$. In order to use multiple edge features, we learn to calculate the scalar $e_{vu}$ from a feature vector $\boldsymbol{e}_{vu}$ using a single-layer perceptron.

\textbf{GraphLSTM}\,\,\, LSTMs \cite{hochreiter1997long} are a class of recurrent neural architectures which use a sequence of gates to keep track of both short- and long-term dependencies in the data. Each of these gates usually applies weights via matrix multiplication; however, we instead replace each of these linear transformations with a graph convolution operation, specifically GraphSAGE, in order to maintain the spatial structure of our inputs and jointly model spatio-temporal dependencies. We refer to this modification of the LSTM architecture as GraphLSTM.

\textbf{Skip Connection}\,\,\, The overall structure of our model (see Figure \ref{fig:arch}) consists of an initial GraphSAGE layer with $n=3$, feeding into a GraphLSTM with embedded GraphSAGE, followed by a final GraphSAGE layer also using $n=3$. A multi-layer perceptron with ReLU activation is applied to finally generate a prediction. Rather than being fed only the output of the GraphLSTM, the final GraphSAGE layer recieves a concatenation of the output of the first GraphSAGE layer and the LSTM. This skip connection is an important contribution of our work, as it is found to speed stabilization of the model and discourage underfitting.

\textbf{Refinements}\,\,\, In addition to the presented structure, in both GraphSAGE layers, we make use of common ML techniques including: 1) dropout, which randomly disables nodes in the graph in order to discourage overfitting and increase model robustness \cite{srivastava2014}, and 2) residual connections, which concatenate raw inputs to posterior layers in order to address the vanishing gradient problem \cite{he2016}.

\textbf{Loss Function}\,\,\, The specific loss function we use is mean absolute scaled error (MASE). To train the model we look at the per-person error, i.e., we aggregate all geographical regions and consider the absolute number of COVID-19 cases. Specifically,
\begin{equation}
\mathcal{L}_t = \frac{|\sum_{i=0}^{N-1}{\hat{X}_{ti}-X_{ti}}|}{\sum_{i=0}^{N-1}{X_{ti}}}
\end{equation}
where $N$ is the total number of geographical regions considered.

\begin{table}[t]
\caption{Per-person and per-country MASE on the test data for our model and each of the 5 compared approaches, including Lag prediction.}
\label{tab:comparison}
\vskip 0.15in
\begin{center}
\begin{small}
\begin{sc}
\begin{tabular}{lcccr}
\toprule
Model & Per-person & Per-country \\
\midrule
Our Model    & \textbf{0.10} & \textbf{0.27} \\
GConvLSTM    & 0.83 & 2.15 \\
GConvGRU     & 0.78 & 1.59 \\
DCRNN        & 0.83 & 2.15 \\
GCLSTM       & 0.83 & 1.66 \\
Lag          & 0.13 & 0.30 \\
\bottomrule
\end{tabular}
\end{sc}
\end{small}
\end{center}
\vskip -0.1in
\end{table}

\begin{figure*}[t]
\centering
\includegraphics[width=.95\linewidth]{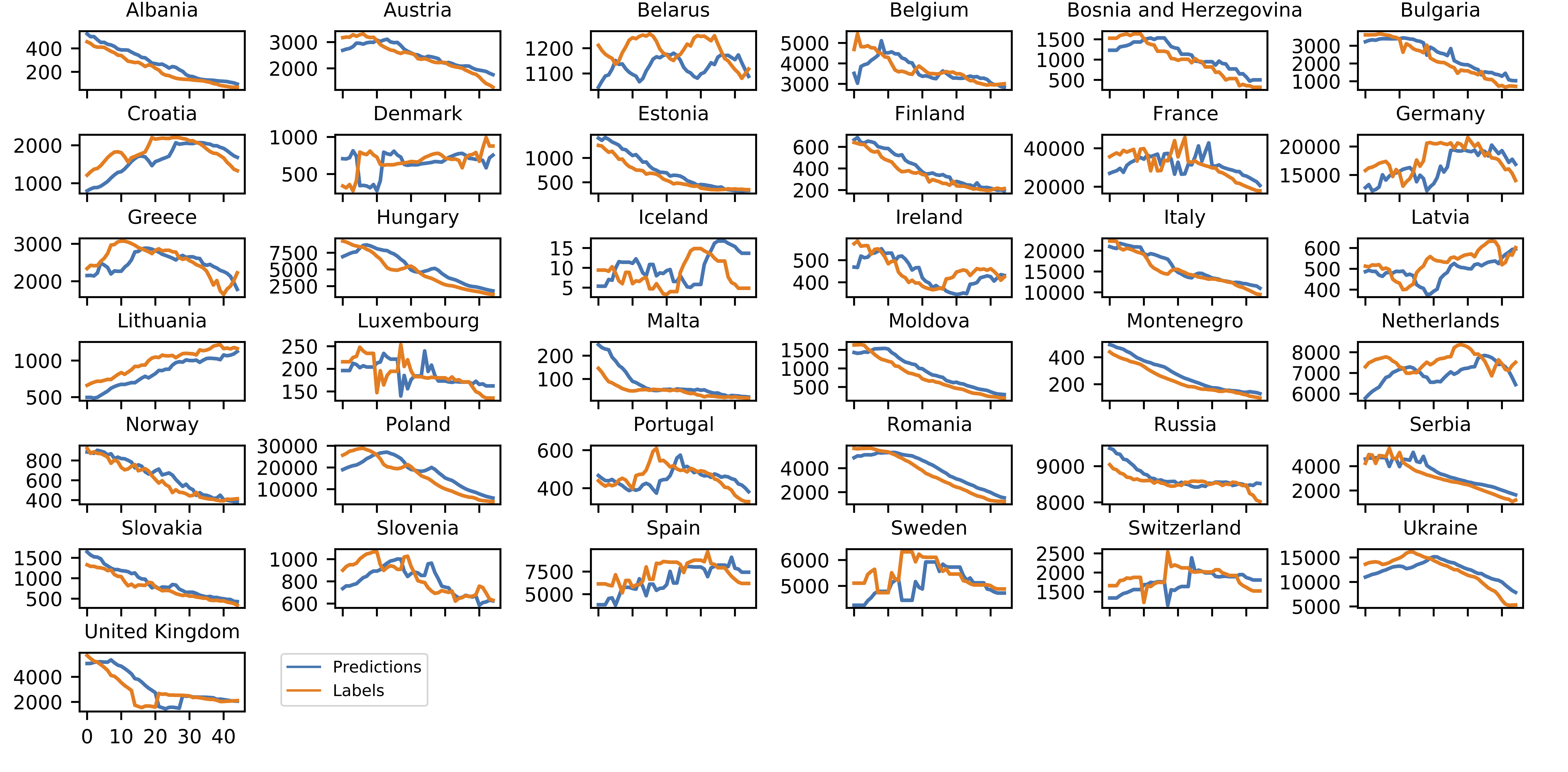}
\caption{Predictions and labels for the test data, consisting of new COVID-19 cases in 37 European countries for the period 22 Mar. 2021 - 9 May 2021 (48 days).}
\label{fig:gridplot}
\end{figure*}

\section{Results}

We now apply our model\footnote{A repository with our code can be found at: \href{https://github.com/jjgarau/GNND}{https://github.com/jjgarau/GNND}} to COVID-19 forecasting; we make use of \cite{rozemberczki2021pytorch}. We start with the dataset and training procedure, and then discuss the results.

\textbf{Dataset and Training}\,\,\, Motivated by the dynamics of the virus in Europe \cite{dye2020scale,saglietto2020covid}, we select a use case consisting of daily new COVID-19 cases for 37 European nations ($N=37$) with more than 100k inhabitants. We use new COVID-19 cases data from \cite{dong2020interactive} for the period 24 Jan. 2020 - 9 May 2021. We use the first 377 days as training data, the following 47 days as validation data, and the last 48 days as test data. Our task consists of predicting the smoothed number\footnote{Moving average over one week} of new cases one week ahead ($M=7$). We train our model by individually passing the data from each of 21 consecutive days ($L=21$), beginning 28 days prior to our prediction target, and use the output of the 21$^{\text{st}}$ day as our prediction. The edges connecting countries are chosen based on proximity, each node being assigned to its 3 nearest neighbors, as determined by geodesic distance between landmass centroids. We use a single edge feature ($K_W = 1$), consisting of the social connectivity score between two incident nations according to \cite{facebookSCI}. 

\textbf{Baseline Comparison}\,\,\, The results of running our model on the aforementioned data are shown in Figure \ref{fig:gridplot}. We compare our approach with four state-of-the-art models which have been shown effective in similar spatial time series forecasting problems: GConvLSTM \cite{seo2018}, GConvGRU \cite{seo2018}, DCRNN \cite{li2017}, and GCLSTM \cite{chen2018}. We find that ours displays significant improvements for the task of COVID-19 prediction. We also compare against lag, which uses the last input date as its prediction. This mechanism has strong results, and though our model learns a similar lagged behaviour, it yields a 20\% reduction in loss over pure lag. Specific numbers are presented in Table \ref{tab:comparison} for both our loss function, per-person MASE, and per-country MASE, which is instead calculated such that each nation weights equally regardless of its total number of cases.

\begin{figure}[t]
\centering
\includegraphics[width=.9\linewidth]{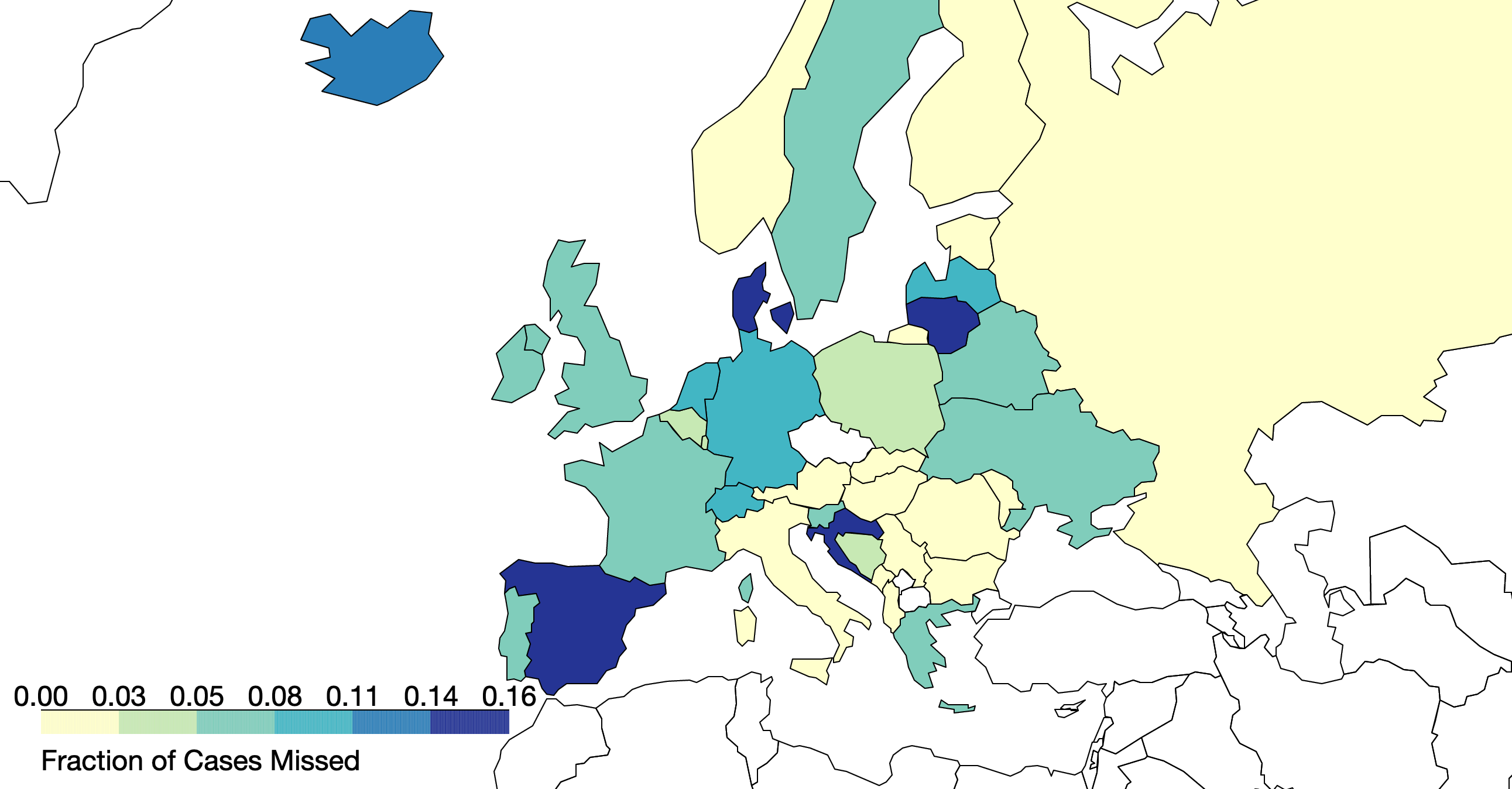}
\caption{Fraction of missed cases on the 48 days of test data.}
\label{fig:missed_cases}
\end{figure}

\textbf{Application to Policy-making}\,\,\, Both of the aforementioned metrics are useful for policy-making: per-person MASE in cases of international policy-making, such as that done by the EU; and per-country MASE in the case where countries must put national policies in place. To better understand the real impact of our predictions, we also present an illustration of the fraction of missed cases (i.e., undercounting) in each country (Figure \ref{fig:missed_cases}) when using our model on the 48 days of test data. Although time series models are usually trained with symmetric loss functions such as MASE or MSE, their real-world impact might be asymmetric, as in the case of COVID-19, where underprediction might hamper preparedness. In our case, the fraction of missed cases ranges from 0 to 0.16, and stays below 0.1 for the majority of the countries.

\textbf{Skip Connection Ablation Study}\,\,\, To validate the efficacy of our proposed skip connection, we perform an ablation test, finding that the altered model achieves an inferior result of 0.76 for per-person MASE (1.68 per-country MASE). We conclude that the skip connection plays a critical role in learning the spatio-temporal patterns present in the data.


\section{Conclusion and Future Work}

In this paper, we presented a model which built upon existing approaches to COVID-19 prediction by further integrating LSTMs and GNNs. The outputs of our model could prove useful to policy-makers attempting to take preemptive action by giving them improved knowledge of future evolution of the pandemic. Our model embeds the GraphSAGE graph convolution operator in place of linear transformations within the gates of an LSTM to create GraphLSTM, a module capable of jointly capturing spatio-temporal patterns. We also propose a skip connection to solve common challenges of time series modeling, which is shown to be an important addition to the model. Our work also presents a solution useful beyond the application of epidemiological prediction. Numerous problems similarly rely on a graph structure with node features changing in time. Future work may also further capitalize on choice of data—while we used only single edge and node features, there are certainly other relevant factors potentially including hospital capacity, poverty rates, and age demographics.

%



\bibliography{library}

\begin{thebibliography}{32}
\providecommand{\natexlab}[1]{#1}
\providecommand{\url}[1]{\texttt{#1}}
\expandafter\ifx\csname urlstyle\endcsname\relax
  \providecommand{\doi}[1]{doi: #1}\else
  \providecommand{\doi}{doi: \begingroup \urlstyle{rm}\Url}\fi

\bibitem[Alazab et~al.(2020)Alazab, Awajan, Mesleh, Abraham, Jatana, and
  Alhyari]{alazab2020covid}
Alazab, M., Awajan, A., Mesleh, A., Abraham, A., Jatana, V., and Alhyari, S.
\newblock Covid-19 prediction and detection using deep learning.
\newblock \emph{International Journal of Computer Information Systems and
  Industrial Management Applications}, 12:\penalty0 168--181, 2020.

\bibitem[Alimadadi et~al.(2020)Alimadadi, Aryal, Manandhar, Munroe, Joe, and
  Cheng]{Alimadadi2020}
Alimadadi, A., Aryal, S., Manandhar, I., Munroe, P.~B., Joe, B., and Cheng, X.
\newblock Artificial intelligence and machine learning to fight covid-19.
\newblock \emph{Physiological Genomics}, 52\penalty0 (4):\penalty0 200--202,
  2020.

\bibitem[Arora et~al.(2020)Arora, Kumar, and Panigrahi]{arora2020}
Arora, P., Kumar, H., and Panigrahi, B.~K.
\newblock Prediction and analysis of covid-19 positive cases using deep
  learning models: A descriptive case study of india.
\newblock \emph{Chaos, Solitons \& Fractals}, 139:\penalty0 110017, 2020.
\newblock ISSN 0960-0779.

\bibitem[Bailey et~al.(1975)]{bailey1975mathematical}
Bailey, N.~T. et~al.
\newblock \emph{The mathematical theory of infectious diseases and its
  applications}.
\newblock Charles Griffin \& Company Ltd, 5a Crendon Street, High Wycombe,
  Bucks HP13 6LE., 1975.

\bibitem[Barstugan et~al.(2020)Barstugan, Ozkaya, and Ozturk]{Barstugan2020}
Barstugan, M., Ozkaya, U., and Ozturk, S.
\newblock {Coronavirus (COVID-19) Classification using CT Images by Machine
  Learning Methods}.
\newblock \emph{arXiv preprint arXiv:2003.09424}, mar 2020.

\bibitem[Buckee et~al.(2020)Buckee, Balsari, Chan, Crosas, Dominici, Gasser,
  Grad, Grenfell, Halloran, Kraemer, Lipsitch, Metcalf, Meyers, Perkins,
  Santillana, Scarpino, Viboud, Wesolowski, and Schroeder]{Sills145}
Buckee, C.~O., Balsari, S., Chan, J., Crosas, M., Dominici, F., Gasser, U.,
  Grad, Y.~H., Grenfell, B., Halloran, M.~E., Kraemer, M. U.~G., Lipsitch, M.,
  Metcalf, C. J.~E., Meyers, L.~A., Perkins, T.~A., Santillana, M., Scarpino,
  S.~V., Viboud, C., Wesolowski, A., and Schroeder, A.
\newblock Aggregated mobility data could help fight covid-19.
\newblock \emph{Science}, 368\penalty0 (6487):\penalty0 145--146, 2020.
\newblock ISSN 0036-8075.

\bibitem[Cao et~al.(2021)Cao, Wang, Duan, Zhang, Zhu, Huang, Tong, Xu, Bai,
  Tong, et~al.]{cao2021}
Cao, D., Wang, Y., Duan, J., Zhang, C., Zhu, X., Huang, C., Tong, Y., Xu, B.,
  Bai, J., Tong, J., et~al.
\newblock Spectral temporal graph neural network for multivariate time-series
  forecasting.
\newblock \emph{arXiv preprint arXiv:2103.07719}, 2021.

\bibitem[Chen et~al.(2018)Chen, Xu, Wu, and Zheng]{chen2018}
Chen, J., Xu, X., Wu, Y., and Zheng, H.
\newblock Gc-lstm: Graph convolution embedded lstm for dynamic link prediction.
\newblock \emph{arXiv preprint arXiv:1812.04206}, 2018.

\bibitem[Chimmula \& Zhang(2020)Chimmula and Zhang]{chimmula2020}
Chimmula, V. K.~R. and Zhang, L.
\newblock Time series forecasting of covid-19 transmission in canada using lstm
  networks.
\newblock \emph{Chaos, Solitons \& Fractals}, 135:\penalty0 109864, 2020.
\newblock ISSN 0960-0779.

\bibitem[Defferrard et~al.(2016)Defferrard, Bresson, and
  Vandergheynst]{defferrard2016}
Defferrard, M., Bresson, X., and Vandergheynst, P.
\newblock Convolutional neural networks on graphs with fast localized spectral
  filtering.
\newblock \emph{arXiv preprint arXiv:1606.09375}, 2016.

\bibitem[Dong et~al.(2020)Dong, Du, and Gardner]{dong2020interactive}
Dong, E., Du, H., and Gardner, L.
\newblock An interactive web-based dashboard to track covid-19 in real time.
\newblock \emph{The Lancet infectious diseases}, 20\penalty0 (5):\penalty0
  533--534, 2020.

\bibitem[Dye et~al.(2020)Dye, Cheng, Dagpunar, and Williams]{dye2020scale}
Dye, C., Cheng, R.~C., Dagpunar, J.~S., and Williams, B.~G.
\newblock The scale and dynamics of covid-19 epidemics across europe.
\newblock \emph{Royal Society open science}, 7\penalty0 (11):\penalty0 201726,
  2020.

\bibitem[{Facebook}(2021)]{facebookSCI}
{Facebook}.
\newblock {Social Connectedness Index}, 2021.
\newblock URL
  \url{https://dataforgood.fb.com/tools/social-connectedness-index/}.
\newblock Accessed: 8 May 2021.

\bibitem[Fauci et~al.(2020)Fauci, Lane, and Redfield]{Fauci2020}
Fauci, A.~S., Lane, H.~C., and Redfield, R.~R.
\newblock Covid-19 — navigating the uncharted.
\newblock \emph{New England Journal of Medicine}, 382\penalty0 (13):\penalty0
  1268--1269, 2020.

\bibitem[Fritz et~al.(2021)Fritz, Dorigatti, and R{\"u}gamer]{fritz2021}
Fritz, C., Dorigatti, E., and R{\"u}gamer, D.
\newblock Combining graph neural networks and spatio-temporal disease models to
  predict covid-19 cases in germany.
\newblock \emph{arXiv preprint arXiv:2101.00661}, 2021.

\bibitem[Hamilton et~al.(2017)Hamilton, Ying, and Leskovec]{hamilton2017}
Hamilton, W.~L., Ying, R., and Leskovec, J.
\newblock Inductive representation learning on large graphs.
\newblock \emph{arXiv preprint arXiv:1706.02216}, 2017.

\bibitem[He et~al.(2016)He, Zhang, Ren, and Sun]{he2016}
He, K., Zhang, X., Ren, S., and Sun, J.
\newblock Deep residual learning for image recognition.
\newblock In \emph{Proceedings of the IEEE conference on computer vision and
  pattern recognition}, pp.\  770--778, 2016.

\bibitem[Hochreiter \& Schmidhuber(1997)Hochreiter and
  Schmidhuber]{hochreiter1997long}
Hochreiter, S. and Schmidhuber, J.
\newblock Long short-term memory.
\newblock \emph{Neural computation}, 9\penalty0 (8):\penalty0 1735--1780, 1997.

\bibitem[La~Gatta et~al.(2021)La~Gatta, Moscato, Postiglione, and
  Sperlí]{LaGatta2021}
La~Gatta, V., Moscato, V., Postiglione, M., and Sperlí, G.
\newblock An epidemiological neural network exploiting dynamic graph structured
  data applied to the covid-19 outbreak.
\newblock \emph{IEEE Transactions on Big Data}, 7\penalty0 (1):\penalty0
  45--55, 2021.

\bibitem[Li et~al.(2020)Li, Guan, Wu, Wang, Zhou, Tong, Ren, Leung, Lau, Wong,
  Xing, Xiang, Wu, Li, Chen, Li, Liu, Zhao, Liu, Tu, Chen, Jin, Yang, Wang,
  Zhou, Wang, Liu, Luo, Liu, Shao, Li, Tao, Yang, Deng, Liu, Ma, Zhang, Shi,
  Lam, Wu, Gao, Cowling, Yang, Leung, and Feng]{Li2020}
Li, Q., Guan, X., Wu, P., Wang, X., Zhou, L., Tong, Y., Ren, R., Leung, K.~S.,
  Lau, E.~H., Wong, J.~Y., Xing, X., Xiang, N., Wu, Y., Li, C., Chen, Q., Li,
  D., Liu, T., Zhao, J., Liu, M., Tu, W., Chen, C., Jin, L., Yang, R., Wang,
  Q., Zhou, S., Wang, R., Liu, H., Luo, Y., Liu, Y., Shao, G., Li, H., Tao, Z.,
  Yang, Y., Deng, Z., Liu, B., Ma, Z., Zhang, Y., Shi, G., Lam, T.~T., Wu,
  J.~T., Gao, G.~F., Cowling, B.~J., Yang, B., Leung, G.~M., and Feng, Z.
\newblock Early transmission dynamics in wuhan, china, of novel
  coronavirus–infected pneumonia.
\newblock \emph{New England Journal of Medicine}, 382\penalty0 (13):\penalty0
  1199--1207, 2020.

\bibitem[Li et~al.(2017)Li, Yu, Shahabi, and Liu]{li2017}
Li, Y., Yu, R., Shahabi, C., and Liu, Y.
\newblock Diffusion convolutional recurrent neural network: Data-driven traffic
  forecasting.
\newblock \emph{arXiv preprint arXiv:1707.01926}, 2017.

\bibitem[Rozemberczki et~al.(2021)Rozemberczki, Scherer, He, Panagopoulos,
  Riedel, Astefanoaei, Kiss, Beres, Lopez, Collignon, and
  Sarkar]{rozemberczki2021pytorch}
Rozemberczki, B., Scherer, P., He, Y., Panagopoulos, G., Riedel, A.,
  Astefanoaei, M., Kiss, O., Beres, F., Lopez, G., Collignon, N., and Sarkar,
  R.
\newblock {PyTorch Geometric Temporal: Spatiotemporal Signal Processing with
  Neural Machine Learning Models}, 2021.

\bibitem[Saglietto et~al.(2020)Saglietto, D’Ascenzo, Zoccai, and
  De~Ferrari]{saglietto2020covid}
Saglietto, A., D’Ascenzo, F., Zoccai, G.~B., and De~Ferrari, G.~M.
\newblock Covid-19 in europe: the italian lesson.
\newblock \emph{Lancet}, 395\penalty0 (10230):\penalty0 1110--1111, 2020.

\bibitem[Santosh(2020)]{santosh2020}
Santosh, K.
\newblock Covid-19 prediction models and unexploited data.
\newblock \emph{Journal of medical systems}, 44\penalty0 (9):\penalty0 1--4,
  2020.

\bibitem[Scarselli et~al.(2008)Scarselli, Gori, Tsoi, Hagenbuchner, and
  Monfardini]{scarselli2008graph}
Scarselli, F., Gori, M., Tsoi, A.~C., Hagenbuchner, M., and Monfardini, G.
\newblock The graph neural network model.
\newblock \emph{IEEE transactions on neural networks}, 20\penalty0
  (1):\penalty0 61--80, 2008.

\bibitem[Seo et~al.(2018)Seo, Defferrard, Vandergheynst, and Bresson]{seo2018}
Seo, Y., Defferrard, M., Vandergheynst, P., and Bresson, X.
\newblock Structured sequence modeling with graph convolutional recurrent
  networks.
\newblock In \emph{International Conference on Neural Information Processing},
  pp.\  362--373. Springer, 2018.

\bibitem[Shahid et~al.(2020)Shahid, Zameer, and Muneeb]{shahid2020}
Shahid, F., Zameer, A., and Muneeb, M.
\newblock Predictions for covid-19 with deep learning models of lstm, gru and
  bi-lstm.
\newblock \emph{Chaos, Solitons \& Fractals}, 140:\penalty0 110212, 2020.
\newblock ISSN 0960-0779.

\bibitem[Srivastava et~al.(2014)Srivastava, Hinton, Krizhevsky, Sutskever, and
  Salakhutdinov]{srivastava2014}
Srivastava, N., Hinton, G., Krizhevsky, A., Sutskever, I., and Salakhutdinov,
  R.
\newblock Dropout: a simple way to prevent neural networks from overfitting.
\newblock \emph{The journal of machine learning research}, 15\penalty0
  (1):\penalty0 1929--1958, 2014.

\bibitem[{United Nations}(2020)]{UnitedNations}
{United Nations}.
\newblock {COVID-19 to slash global economic output by \$8.5 trillion over next
  two years}, 2020.
\newblock URL
  \url{https://www.un.org/development/desa/en/news/policy/wesp-mid-2020-report.html}.
\newblock Accessed: 8 May 2021.

\bibitem[Velavan \& Meyer(2020)Velavan and Meyer]{velavan2020covid}
Velavan, T.~P. and Meyer, C.~G.
\newblock The covid-19 epidemic.
\newblock \emph{Tropical medicine \& international health}, 25\penalty0
  (3):\penalty0 278, 2020.

\bibitem[Wang et~al.(2021)Wang, Govindaraj, Górriz, Zhang, and
  Zhang]{WANG2021208}
Wang, S.-H., Govindaraj, V.~V., Górriz, J.~M., Zhang, X., and Zhang, Y.-D.
\newblock Covid-19 classification by fgcnet with deep feature fusion from graph
  convolutional network and convolutional neural network.
\newblock \emph{Information Fusion}, 67:\penalty0 208--229, 2021.
\newblock ISSN 1566-2535.

\bibitem[{World Health Organization}(2021)]{WorldHealthOrganization}
{World Health Organization}.
\newblock {WHO Coronavirus (COVID-19) Dashboard}, 2021.
\newblock URL \url{https://covid19.who.int/}.
\newblock Accessed: 8 May 2021.

\end{thebibliography}
\bibliographystyle{icml2021}


\end{document}